\documentclass[10pt,twocolumn,letterpaper]{article}

\usepackage[pagenumbers]{wacv} % To force page numbers, e.g. for an arXiv version

\usepackage{graphicx}
\usepackage{amsmath}
\usepackage{amssymb}
\usepackage{booktabs}
\usepackage{tabularx}
\usepackage{comment}
\usepackage{algorithm2e}
\usepackage{caption}
\usepackage{arydshln}

\usepackage[pagebackref,breaklinks,colorlinks]{hyperref}

\usepackage[capitalize]{cleveref}
\crefname{section}{Sec.}{Secs.}
\Crefname{section}{Section}{Sections}
\Crefname{table}{Table}{Tables}
\crefname{table}{Tab.}{Tabs.}

\def\wacvPaperID{1132} % *** Enter the WACV Paper ID here
\def\confName{WACV}
\def\confYear{2025}

\begin{document}

\title{Inverse Problems with Diffusion Models: A MAP Estimation Perspective}

\author{
Sai bharath chandra Gutha\ \ \ \ \ \ \ \ Ricardo Vinuesa\ \ \ \ \ \ \ \ Hossein Azizpour\\
{\tt\small sbcgutha@kth.se, rvinuesa@mech.kth.se, azizpour@kth.se}\\
{KTH, Sweden}\\\\
%\\
%{\tt\small sbcgutha@kth.se}
%\and
%Ricardo Vinuesa\\
%{\tt\small rvinuesa@mech.kth.se}
%\and
%Hossein Azizpour\\
%{\tt\small azizpour@kth.se}\\
%{KTH, Sweden}
}
\maketitle
%%%%%%%%% ABSTRACT
\begin{abstract}
Inverse problems have many applications in science and engineering. In Computer vision, several image restoration tasks such as inpainting, deblurring, and super-resolution can be formally modeled as inverse problems. Recently, methods have been developed for solving inverse problems that only leverage a pre-trained unconditional diffusion model and do not require additional task-specific training. In such methods, however, the inherent intractability of determining the conditional score function during the reverse diffusion process poses a real challenge, leaving the methods to settle with an approximation instead, which affects their performance in practice. Here, we propose a MAP estimation framework to model the reverse conditional generation process of a continuous time diffusion model as an optimization process of the underlying MAP objective, whose gradient term is tractable. In theory, the proposed framework can be applied to solve general inverse problems using gradient-based optimization methods. However, given the highly non-convex nature of the loss objective, finding a perfect gradient-based optimization algorithm can be quite challenging, nevertheless, our framework offers several potential research directions. We use our proposed formulation to develop empirically effective algorithms for image restoration. We validate our proposed algorithms with extensive experiments over multiple datasets across several restoration tasks.       
%  Inverse problems have many applications in Science and Engineering. In Computer vision, several Image restoration tasks such as inpainting, deblurring, and super-resolution can be formally modeled as inverse problems. Recently, methods have been developed for solving inverse problems that leverage a pre-trained unconditional diffusion model and importantly do not require any additional task-specific training. In such methods, however, the inherent intractability of determining the conditional score function in the reverse diffusion process poses a real challenge, leaving the methods to settle with an approximation instead, which affects their performance in practice. Here, we propose a MAP estimation framework to model the reverse conditional generation process of a continuous time diffusion model as an optimization process of the underlying MAP objective, whose gradient term is tractable. In theory, the proposed framework can be applied to solve Inverse problems using gradient-based optimization methods. However, given the highly non-convex nature of the loss objective, finding a perfect gradient-based optimization algorithm can be quite challenging, nevertheless, our framework offers a novel direction for improvement and several potential research directions. We use our proposed formulation, for solving noiseless and noisy image inpainting tasks by developing empirically effective algorithms that require few or no hyperparameter tuning. We validate the effectiveness of our approach with extensive experiments on Image Inpainting across diverse mask settings.     
\end{abstract}

%%%%%%%%% BODY TEXT
\section{Introduction}
\label{sec:introduction}
Inverse problems are ubiquitous in science and engineering with a wide range of downstream applications ~\cite{ipse, bora-compressed}. In Computer vision, several image restoration tasks such as inpainting, deblurring, super-resolution, and more, can be formally modeled as inverse problems. In an inverse problem, characterized by \cref{eqn:invprob}, $y \in \mathbb{R}^m$ is a (potentially noisy) observation of the original data $x \in \mathbb{R}^n$, and $\eta$ is a random variable denoting i.i.d. noise, typically assumed to be Gaussian with a known variance i.e $\eta \sim \mathcal{N}(0,\sigma^2_y \mathbb{I})$, and the task is to infer the original data $x$ given the observation $y$. The function $\mathcal{A}: \mathbb{R}^n \to \mathbb{R}^m$ is known as the forward operator, and typically $n \gg m$, indicating that the observation $y \in \mathbb{R}^m$ corresponds to a severely degraded signal, from which one needs to recover the original signal $x \in \mathbb{R}^n$, which makes the task highly challenging. For linear inverse problems, $\mathcal{A}$ denotes a linear mapping and can be substituted with a matrix $H \in \mathbb{R}^{m\times n}$. 
\begin{equation}\label{eqn:invprob}
    y = \mathcal{A}(x) + \eta
\end{equation}
Several conventional approaches for solving inverse problems exist~\cite{acta}. These include approaches based on functional-analytic, probabilistic, data-driven methods, and more. Recently, Deep Learning (DL) based methods have been applied to solve inverse problems and have shown great success. In a Bayesian framework, solving an inverse problem naturally corresponds to estimating the posterior $P(x|y)$. Typical DL-based approaches for solving inverse problems fall into two categories. \textbf{1.}\label{text:cat1} Methods that directly learn the posterior $P(x|y)$ via conditional generative models~\cite{glide,palette}, and \textbf{2.}\label{text:cat2} Methods that learn $P(x)$ via an unconditional generative model and use it to infer $P(x|y)$ ~\cite{pgdm,dps,dmplug,ddrm}. Methods of the former category require task-specific training, i.e. training with a dataset of pairs $(x,y)$, where the degradation $y$ is computed using $x$ and a task-specific forward operator $\mathcal{A}$. This limits the out-of-the-box applicability of the model to a different task (different forward operator). On the contrary, methods of the latter category train an unconditional generative model to learn $P(x)$, and this training is task-independent since it only needs a dataset of original data samples $x$. These methods then use the trained model for $P(x)$ and since $P(y|x)$ is tractable (i.e. from \cref{eqn:invprob}, $P(y|x) = \mathcal{N}(\mathcal{A}(x), \sigma^2_y \mathbb{I})$), utilizing the Bayes rule, they infer the posterior $P(x|y) \propto P(y|x) P(x)$.\\

Several choices for Deep Generative Models (DGMs) exist, each with its advantages and disadvantages. There have been approaches using Generative Adversarial Networks (GAN)~\cite{gans} and Normalizing Flow (NF)~\cite{nf} based DGMs for solving inverse problems, with more recent methods focusing on Diffusion models~\cite{sohl,ddpm,song-sde}, owing to their state-of-the-art performance in several vision-based generative tasks. This work focuses on methods that use a pre-trained unconditional diffusion model as the prior $P(x)$ and infer the posterior $P(x|y)$ for solving inverse problems. \cref{sec:background} provides some background on diffusion models and related works that use unconditional diffusion models for solving inverse problems and their inherent limitations. Later in ~\cref{sec:method}, we propose our Maximum A Posteriori (MAP) estimation framework for continuous-time diffusion models and discuss the practical implementation. Specifically, we propose a novel MAP formulation that employs a reparameterization based on consistency models~\cite{song-consistency} to model the reverse conditional diffusion process as MAP optimization. In ~\cref{sec:imageinpainting}, we use our proposed framework to develop empirically effective algorithms for image restoration, and in ~\cref{sec:experiments}, we validate the algorithms with extensive experiments on deblurring, super-resolution, and image inpainting. In ~\cref{sec:discussion}, we present a brief discussion before concluding our findings in ~\cref{sec:conclusion}.
%The core requirement in several of these tasks is the need to model conditional generation i.e. generating input based on a given observation or condition. While the former can be achieved by directly modeling and training a conditional Generative Diffusion model, this limits the out-of-the-box applicability of the trained model to a different task since its training needs to be task-specific. 
%------------------------------------------------------------------------
\section{Background}
\label{sec:background}
Given a dataset of samples $\mathcal{D} = {\{x_{i}\}}_{i=1}^{N}$, where each $x_i$ is an i.i.d. sample drawn from an unknown data distribution $P_{data}(x)$, a generative model learns to approximate $P_{data}$ from the samples in $\mathcal{D}$.

\subsection{Diffusion Models}
\label{subsec:diffusionmodels}
Diffusion models ~\cite{sohl,ddpm,song-sde} are a recent family of generative models. The methodology involves simulating a stochastic process $\{x(t)\}_{t=0}^{T}$ described by a Stochastic Differential Equation (SDE) such as \cref{eqn:forwardsde}, where $t \in [0,T]$ is a continuous time variable, $x(0) \sim P_0 = P_{data}$ is the data distribution for which we have a dataset $\mathcal{D}$ of samples, $x(T) \sim P_T$ is a tractable prior distribution. The functions $f(\cdot,t):\mathbb{R}^n \to \mathbb{R}^n$ and $g(\cdot): \mathbb{R} \to \mathbb{R}$ are called the drift and diffusion coefficients of $x(t)$ respectively, and $dw$ denotes the standard Wiener process. Typically $f$ and $g$ are chosen in a way that yields a tractable prior $P_T$ which contains no information about $P_0$ (i.e. $P_{data}$). 
\begin{equation}\label{eqn:forwardsde}
    dx = f(x,t)dt + g(t)dw
\end{equation}
\cref{eqn:forwardsde} also describes the "forward process", in which, starting from an initially clean data sample, a small amount of noise is progressively added at each step until it turns into a noisy sample of the prior distribution $P_T$. The backward/reverse process which transforms a noisy sample of $P_T$ into a clean sample of the data distribution is described by the corresponding reverse-SDE in \cref{eqn:reversesde} 
\begin{equation}\label{eqn:reversesde}
    dx = [f(x,t) - g(t)^2 \nabla_{x} \log P_t(x)]dt + g(t) d\bar{w} 
\end{equation}
$d\bar{w}$ denotes the standard Wiener process when $t$ flows backwards from $T$ to $0$ with $dt$ denoting an infinitesimal negative time step. The term $\nabla_{x} \log P_t(x)$ is called the score function of the marginal distribution $P_t(x)$. If we know this score function for each marginal distribution i.e. for all $t$, then one could solve the reverse-SDE in \cref{eqn:reversesde} and generate samples from the data distribution. In general, the score function is not analytically tractable and is hard to estimate, however, one could train a time-indexed neural network model to learn the score function via score matching techniques~\cite{vincent-dsm,song-ssm}. The trained score model $S_{\theta}(x,t)$ can be substituted in place of $\nabla_{x}\log P_t(x)$ in \cref{eqn:reversesde}, and the reverse-SDE can be solved using traditional SDE solvers~\cite{song-sde}.
\begin{equation}\label{eqn:pfode}
    dx = [f(x,t) - \frac{1}{2} g(t)^2 \nabla_x \log P_t(x)]dt
\end{equation}
For the stochastic process described by the SDE in \cref{eqn:forwardsde}, there exists a corresponding Ordinary Differential Equation (ODE) shown in \cref{eqn:pfode}, describing a deterministic process whose trajectories share the same marginal probability densities $\{P(x(t))\}_{t=0}^{T}$ as those simulated by the SDE. This is called the Probability Flow ODE (PF ODE) in the literature~\cite{song-sde}. So equivalently, one could also use ODE solvers to solve \cref{eqn:pfode} in reverse time from $t=T$ until $0$ to generate samples from the data distribution.\\

Hereon, we assume the default choice for drift and diffusion coefficients as $f(x,t) = 0$ and $g(t) = \sqrt{\frac{d\sigma^2(t)}{dt}}$, where $\sigma(t)$ is a monotonically increasing noise schedule from $t=0$ to $T$, with $\sigma(T)$ being very high. This choice of $f$ and $g$ results in a closed form perturbation kernel $P(x(t)|x(0)) = \mathcal{N}(x(0), \sigma^2(t)-\sigma^2(0))$ and $P(x(T)) \approx \mathcal{N}(0, \sigma^2(T))$. There can be multiple design choices for the noise schedule $\sigma(t)$, resulting in various formulations of diffusion models ~\cite{edm}.

\subsection{Solving Inverse problems with Diffusion models}
\label{subsec:invprobwithdiffusion}

As described in \cref{sec:introduction}, solving an inverse problem entails estimation of (or sampling from) the posterior $P(x|y)$, where $y$ is the noisy degradation of the original data sample $x$ (\cref{eqn:invprob}). In the context of solving inverse problems using diffusion models, sampling from the posterior $P(x(0)|y)$ involves conditioning the reverse diffusion process on $y$ which translates to solving the modified reverse-SDE in \cref{eqn:reversesdeposterior}.%, in which the unconditional score function is replaced with the conditional score function. 
\begin{equation}\label{eqn:reversesdeposterior}
    dx = [f(x,t) - g(t)^2 \nabla_{x} \log P_t(x|y)]dt + g(t) d\bar{w} 
\end{equation}
Similar to methods of the first category (\cref{text:cat1}), which directly learn the posterior $P(x|y)$ as part of their training, it is possible to train a conditional diffusion model that learns the conditional score function directly. More specifically, one could learn $S_{\theta}(x,y,t)$ using conditional score matching objectives in place of the usual unconditional score function $S_{\theta}(x,t)$ in \cref{subsec:diffusionmodels}. In this work, we focus on methods of the second category, which only leverage an unconditional diffusion model for $P(x)$, to infer $P(x|y)$. We modify the notations to denote $x(t)$ with $x_t$, $\sigma(t)$ with $\sigma_t$, and $P_t(x)$ with $P(x_t)$ for the sake of convenience and to be consistent with previous works~\cite{dps,pgdm}.
\begin{equation}\label{eqn:decomp}
    \resizebox{.85\linewidth}{!}{$\nabla_{x_t}\log P(x_t|y) = \nabla_{x_t}\log P(x_t) + \nabla_{x_t}\log P(y|x_t)$} 
\end{equation}
\begin{equation}\label{eqn:intrac}
    \resizebox{.65\linewidth}{!}{$P(y|x_t) = \int_{x_0} P(y|x_0) P(x_0|x_t) dx_0$} 
\end{equation}\\
Solving \cref{eqn:reversesdeposterior} involves estimating the conditional score function $\nabla_{x_t}\log P(x_t|y)$. The pre-trained unconditional diffusion model can be used to estimate $\nabla_{x_t}\log P(x_t)$, however, the term $\nabla_{x_t}\log P(y|x_t)$ becomes intractable (ref \cref{eqn:decomp,eqn:intrac}). At its core, the intractability of $\nabla_{x_t}\log P(y|x_t)$ arises from the fact that $P(x_0|x_t)$ is intractable~\cite{pgdm}, and hence, the conditional score is hard to estimate while only leveraging the unconditional score. 

\subsection{Related works}\label{ssec:relatedworks}
PGDM~\cite{pgdm} approximates $P(x_0|x_t)$ with a Gaussian distribution having mean $\hat{x_t}$ and variance ${r_t}^2$, where $\hat{x_t} = \mathbb{E}(x_0|x_t) = x_t + \sigma^2_t \nabla_{x_t}\log P(x_t)$ (using Tweedie's formula). The standard deviation $r_t$ is a hyperparameter, chosen proportionally to $\sigma_t$. DPS~\cite{dps} approximates $P(y|x_t)$ with the point estimate $P(y|x_0=\hat{x_t})$ and has an almost similar formulation as PGDM up to a constant factor, though the motivation is slightly different. ~\cite{boys} also uses Gaussian approximation but further replace $r_t$ with the covariance matrix $Cov[x_0|x_t] = \sigma^2_t \frac{\partial \hat{x_t}}{\partial x_t}$. Computing this matrix is expensive in practice, so they resort to diagonal and row-sum approximations of the matrix instead. ~\cite{optcov} proposes to find an optimal covariance matrix using learned covariances from the diffusion model. All these methods, however, still make simplified approximations for $P(x_0|x_t)$, which limits their performance in practice, given the complicated and multimodal nature of the true data distribution. Other lines of work ~\cite{ddrm,ddnm,mcg,dds} try to circumvent this term by projecting the intermediate $x_t$ onto the measurement subspace using heuristic approximations.\\

DiffPIR~\cite{diffpir} poses the problem as MAP optimization with data and prior terms and utilizes the HQS~\cite{hqs} algorithm to solve a relaxed problem where the data and the prior terms can be optimized alternatively in a decoupled manner. Specifically, during each reverse diffusion step, this amounts to solving a relaxed MAP objective. DDS~\cite{dds} uses a similar framework but employs subspace projection methods to solve the intermediate MAP objectives at each diffusion step. ZSIR~\cite{zsir} and DMPlug~\cite{dmplug} also pose the problem as MAP optimization, where instead of optimizing for the original data $x_0$ directly, they reparameterize $x_0$ via the initial diffusion noise $x_T$ and solve for the optimal noise $x_T$ instead. From a theoretical perspective, it is unclear how this reparameterization should help. Also, both works ignore the prior term while focusing only on the data term. If the data distribution has full support i.e. $P(x_0) \neq 0$  $\forall x_0 \in \mathbb{R}^n$, ignoring the prior term, especially in case of a noisy measurement $y$ is theoretically unjustified. Our proposed method in this paper is similar to ZSIR and DMPlug from a practical perspective, however, our MAP formulation and algorithms are based on sound theoretical motivation that justifies the reparameterization based on PF ODE and provides new insights into modeling the conditional generation process as MAP optimization. 
%------------------------------------------------------------------------
\section{Our Methodology}\label{sec:method}

\subsection{Background: Consistency Models}\label{ssec:consistency}
Consider the PF ODE described in \cref{eqn:pfode}. The solution trajectories of this ODE are smooth, and map the samples on the data manifold to pure noise. In ~\cite{song-consistency}, a consistency model is defined as the function that maps any point on the PF ODE trajectory to its corresponding origin (initial point on the data manifold). There exist efficient methodologies~\cite{song-consistency-improved} to train these consistency models in practice. Please refer to ~\cite{song-consistency} for a detailed description.
%In \cref{fig:consistency_model}, $f_{\theta}$ denotes the consistency model mapping any two points on the same PF ODE trajectory to the same sample (trajectory origin).
%\begin{figure}
%    \centering
%    \includegraphics[width=\linewidth,keepaspectratio]{images/consistency_model_fig.pdf}
%    \caption{( \centering Fig. from ~\cite{song-consistency}, illustrating the PF ODE trajectories and the consistency model \textcolor{red}{$f_{\theta}(.,.)$} mapping any two points on a PF ODE trajectory to its origin.)}
%    \label{fig:consistency_model}
%\end{figure}

\subsection{Proposed MAP estimation framework}\label{ssec:mapframework}
\begin{equation}\label{eqn:usualmapformulation}
    x^{*}_{0} = \arg \max_{x_0}\ \log P(x_0|y)
\end{equation}
\cref{eqn:usualmapformulation} refers to the usual MAP formulation for solving an inverse problem. Finding an optimal $x^{*}_{0}$ using gradient ascent involves the update step in \cref{eqn:usualmapgd}, with $k$ denoting the $k^{th}$ iterate and $\lambda$ denoting the step size.
\begin{equation}\label{eqn:usualmapgd}
    x^{k+1}_{0} = x^{k}_{0} + \lambda * \nabla_{x_0} \log P(x_0|y)
\end{equation}
The update step requires computing the gradient term $\nabla_{x_0}\log P(x_0|y)=\nabla_{x_0}\log P(y|x_0)+\nabla_{x_0}\log P(x_0)$. The former term is tractable since $P(y|x_0)$ is Gaussian, and the latter is the score function evaluated at $x_0$ and can be replaced with $S_{\theta}(x_0,0)$.
%One could perform gradient ascent to find a sample $x^{*}_{0}$ with high likelihood and consistent with the measurement $y$. 
In practice, $S_{\theta}(x_0,0)$ is only accurate when $x_0$ lies closer to the data manifold and is typically inaccurate for $x_0$ in low-likelihood regions outside the data manifold~\cite{song-gradient}. This makes the score estimate inaccurate in the beginning (when $x_0$ is initialized randomly) and during the gradient ascent updates, since the intermediate $x^{k}_0$ are not constrained to lie on the data manifold. This issue can be avoided when inverse problems are typically solved through the reverse diffusion process (\cref{subsec:invprobwithdiffusion}) which drifts the noisy sample towards the data manifold using $S_{\theta}(x_t,t)$ while simultaneously ensuring measurement consistency. But there the challenge is to estimate $\nabla_{x_t}\log P(y|x_t)$ which is again intractable, as discussed earlier.\\

Here, we present our proposed MAP formulation. Let $z \sim P(x_T) = \mathcal{N}(0, \sigma^2_T\mathbb{I})$, denote a purely noisy sample, and $\mathcal{M}$ denote the data Manifold. The PF ODE trajectory maps $z$ to a sample $x_0 \in \mathcal{M}$, given by $x_0 = f_{\theta}(z,T)$, where $f_{\theta}$ is the consistency model. It is also evident that, $\forall\ x_0 \in \mathcal{M}$, $\exists\ z \sim P(x_T)$ such that $x_0 = f_{\theta}(z,T)$. Hence, the usual MAP formulation in \cref{eqn:usualmapformulation} is equivalent to the proposed MAP formulation in \cref{eqn:proposedmapformulation1,eqn:proposedmapformulation2}.
\begin{equation}\label{eqn:proposedmapformulation1}
    z^{*} = \arg \max_{z}\ \log P(x_0=f_{\theta}(z,T)|y)
\end{equation}
\begin{equation}\label{eqn:proposedmapformulation2}
    x^{*}_{0} = f_{\theta}(z^{*},T)
\end{equation}
With our proposed formulation, we update $z$ with gradient ascent steps as in \cref{eqn:proposedmapgd} for finding $z^{*}$. The update step now requires computing the gradient term $\nabla_{z} \log P(f_{\theta}(z, T)|y)$, which can be reformulated as a vector-Jacobian product (vjp) as shown in \cref{eqn:proposedmapgradient}. The vector in this vjp is the gradient term $\nabla_{x_0}\log P(x_0|y)$ evaluated at $x_0 = f_{\theta}(z,T)$ which lies on the data manifold (by definition of consistency model) and can be accurately evaluated, unlike the previous case. 
\begin{equation}\label{eqn:proposedmapgd}
    z^{k+1} = z^{k} + \lambda * \nabla_{z} \log P(f_{\theta}(z, T)|y)
\end{equation}    
\begin{equation}\label{eqn:proposedmapgradient}
    \resizebox{.85\linewidth}{!}{$\nabla_{z} \log P(f_{\theta}(z, T)|y) = {\left( \frac{\partial f_{\theta}(z,T)}{\partial z} \right)}^{\intercal}\nabla_{x_0}\log P(x_0|y)\bigg\vert_{x_0=f_{\theta}(z,T)}$}
\end{equation}    
Here we provide a high-level overview of a practical implementation using our MAP formulation. In practice, even a consistency model $f_{\theta}$ can benefit from multi-step sampling ~\cite{song-consistency}. Therefore we propose a multi-step gradient ascent scheme called \emph{\textbf{MAP-Gradient-Ascent (MAP-GA)}} as described in \cref{alg:mapgaopseudo}. Note that $\tau$ refers to a time-step schedule with $T = \tau_n > \tau_{n-1} > .. > \tau_1 > \tau_0 = 0$ and $\sigma$ refers to the monotonically increasing noise schedule for $t \in [0,T]$, with $\sigma_0 = 0,$ and $\sigma_T = \infty$ (high value in practice), $y$ is the measurement, $f_{\theta}$ and $S_{\theta}$ are the consistency model and the score model respectively. $num\_iter$ denotes the number of gradient ascent iterations per time step, and $\lambda$ denotes the learning rate. \cref{alg:mapgaopseudo} can be applied to solve any inverse problem effectively, given that we know the optimal hyper-parameters, such as the time-step schedule $\tau$, the learning rate $\lambda$, learning rate schedule, $num\_steps$, etc. However, finding those in practice can be quite challenging.

\RestyleAlgo{ruled}
\begin{algorithm}
\caption{MAP-GA (MAP-Gradient-Ascent)}\label{alg:mapgaopseudo}
%\SetKwData{Left}{left}\SetKwData{This}{this}\SetKwData{Up}{up}
%\SetKwFunction{Union}{Union}\SetKwFunction{FindCompress}{FindCompress}
\SetKwInOut{Input}{input}\SetKwInOut{Output}{output}

\Input{$\tau=(\tau_n,..\tau_1,\tau_0)$, \emph{$f_{\theta}$}, \emph{$S_{\theta}$}, \emph{y}, \emph{num\_iter}, \emph{$\lambda$}, \emph{$\sigma$}}
{$z \sim \mathcal{N}(0,\sigma^2_{\tau_n}\mathbb{I})$}\\
\For{$i\ in\ (n, n-1,..,1)$}
{
$t = \tau_i$\\
\For{$j\ in\ (1, 2,..,num\_iter)$}{
$z = z + \lambda * \nabla_{z} \log P(f_{\theta}(z,t)|y)$\\
}
$\hat{x}_0 = f_{\theta}(z,t)$\\
$z = \mathcal{N}(\hat{x}_0, \sigma^2_{\tau_{i-1}}\mathbb{I})$\\
}
\Output{z}
\end{algorithm}

\subsection{Practical Implementation}\label{ssec:practical}

In practice, to avoid numerical issues and to ensure that the gradient term $\nabla_{x_0}\log P(x_0)$ exists, instead of solving for $x^{*}_0$, 
%as in \cref{eqn:proposedmapformulation1,eqn:proposedmapformulation2}, 
we solve for $x^{*}_{\epsilon} = \arg \max_{x_\epsilon}\ \log P(x_\epsilon|y)$, for a small $\epsilon$ such that $\sigma_\epsilon \approx 0$. Hence we solve the MAP formulation in \cref{eqn:practiacalmapformulation1,eqn:practiacalmapformulation2} in practice. Since, $P(x_\epsilon|x_0) = \mathcal{N}(x_0, \sigma^{2}_{\epsilon}\mathbb{I})$, for very small values of $\sigma_{\epsilon}$, the distinction between $x_0$ and $x_\epsilon$ remain insignificant for all practical purposes. The consistency models in ~\cite{song-consistency} are also learned to map the points on PF ODE trajectory to the corresponding $x_\epsilon$ instead of $x_0$.
\begin{equation}\label{eqn:practiacalmapformulation1}
    z^{*} = \arg \max_{z}\ \log P(x_\epsilon=f_{\theta}(z,T)|y)
\end{equation}
\begin{equation}\label{eqn:practiacalmapformulation2}
    x^{*}_{0} \approx x^{*}_{\epsilon} = f_{\theta}(z^{*},T)
\end{equation}
%\textbf{Computing the Gradient of log Posterior\\}
Here, we describe in detail the computation of the gradient term $\nabla_{z}\log P(f_{\theta}(z,t)|y)$ in practice. From \cref{eqn:solve1,eqn:solve2}, this requires the estimation of the gradient of log-likelihood i.e. $\nabla_{x_\epsilon}\log P(y|x_\epsilon)$ and the gradient of log-prior i.e. $\nabla_{x_\epsilon}\log P(x_\epsilon)$. The terms $P(y|x_\epsilon)$ and $P(x_\epsilon)$ are also referred to as the \textbf{likelihood} and the \textbf{prior} respectively. 
\begin{equation}\label{eqn:solve1}
    \resizebox{.85\linewidth}{!}{$\nabla_{z} \log P(f_{\theta}(z, t)|y) = {\left( \frac{\partial f_{\theta}(z,t)}{\partial z} \right)}^{\intercal}\nabla_{x_\epsilon}\log P(x_\epsilon|y)\bigg\vert_{x_\epsilon=f_{\theta}(z,t)}$}
\end{equation}
\begin{equation}\label{eqn:solve2}
    \resizebox{.85\linewidth}{!}{$\nabla_{x_\epsilon}\log P(x_\epsilon|y)\bigg\vert_{x_\epsilon=f_{\theta}(z,t)} =\biggl\{ \nabla_{x_\epsilon}\log P(y|x_\epsilon) + \nabla_{x_\epsilon}\log P(x_\epsilon)\biggr\}\bigg\vert_{x_\epsilon=f_{\theta}(z,t)}$}
\end{equation}
\textbf{Computing the gradient of log-likelihood\\} $P(y|x_0) = \mathcal{N}(\mathcal{A}(x_0),\sigma^2_y\mathbb{I})$, and $P(x_\epsilon|x_0) = \mathcal{N}(x_0, \sigma^2_\epsilon \mathbb{I})$, given by \cref{eqn:invprob} and the diffusion perturbation kernel respectively. Since $\sigma_{\epsilon} \approx 0$, $P(x_0)\approx P(x_\epsilon)$ and $P(x_0|x_\epsilon) \approx P(x_\epsilon|x_0)$. We can approximate $P(x_0|x_\epsilon) = \mathcal{N}(x_\epsilon, \sigma^2_\epsilon \mathbb{I})$, and for a linear forward operator i.e. $\mathcal{A} = H \in \mathbb{R}^{m\times n}$, we can derive using \cref{eqn:intrac}, $P(y|x_\epsilon) = \mathcal{N}(Hx_\epsilon, \sigma^2_y\mathbb{I} + \sigma^2_\epsilon HH^{\intercal})$. For non-linear $\mathcal{A}$, similar approximations can be made by linearizing it around $x_\epsilon$. Given the tractable form of the likelihood term above, the gradient of the log-likelihood is apparent.\\
\\
\textbf{Computing the gradient of log-prior\\}
The gradient of the log prior i.e. $\nabla_{x_\epsilon} \log P(x_\epsilon)$ is essentially the score function evaluated at $x_\epsilon$. Given a score function $S_{\theta}(x,t)$, learned by the unconditional diffusion model, $\nabla_{x_\epsilon} \log P(x_\epsilon) = S_{\theta}(x_\epsilon,\epsilon)$. Learning the score function is equivalent to learning a denoiser, and vice-versa~\cite{edm}. Hereon, we denote the unconditional diffusion model as learning the denoiser $D_{\theta}(x,t)$, from which the score function can be computed as \scalebox{0.75}{%
$\nabla_{x_t} \log P(x_t)$}$=\frac{D_{\theta}(x_t,t) - x_t}{\sigma^2_t}$.
%\\

\section{Image restoration with MAP-GA}\label{sec:imageinpainting}

Several image restoration tasks such as inpainting, deblurring, and super-resolution can be modeled as linear inverse problems. For example, in image inpainting~\cite{inpainting1,inpainting2}, given a masked image (and the corresponding binary mask), the goal is to recover (reconstruct) the missing pixels of the masked image. Let $x\in \mathbb{R}^n$ denote the original image, $y \in \mathbb{R}^m$ denote a masked image with only visible pixels ($m \leq n$) and $H \in \mathbb{R}^{m\times n}$ denotes a corresponding linear forward operator for a given mask. The inpainting problem is characterized by $y = Hx + \eta$, where $\eta \sim \mathcal{N}(0,\sigma^2_y \mathbb{I})$. Note that for inpainting, a given mask defines $H$ with a defined structure, where the rows of $H$ are one-hot and are orthogonal i.e. $HH^{\intercal}=\mathbb{I}_{m\times m}$. Other image restoration problems such as deblurring and super-resolution can be modeled accordingly with their corresponding forward operators.\\
%In this work, to show the effectiveness of our proposed MAP framework in practice, we consider the tasks of deblurring, super-resolution, and image inpainting with diverse masks. 

We expand \cref{alg:mapgaopseudo} for image restoration and include all the specific details in \cref{alg:mapgainpaint}. The core term in the algorithm that needs to be evaluated is $\nabla_{z}\log P(f_{\theta}(z,t)|y)$, which involves the estimation of gradients of the log-likelihood and the log-prior terms (\cref{eqn:solve1,eqn:solve2}). In the algorithm, we make a choice (indicated by the $use\_prior$ keyword) of retaining or dropping the gradient of the log-prior. Dropping the prior term implies a choice of uniform prior, and the algorithm now optimizes for the maximum likelihood estimate instead of the MAP estimate, by considering any sample (consistent with our measurement) on the data manifold to be equally good. This is also the setting considered in the concurrent works ~\cite{zsir,dmplug}.\\

Note that the algorithm makes use of both the consistency model ($C_\theta$) and the denoiser ($D_{\theta}$), which makes our method more demanding compared to other methods that only use the denoiser. We make an argument as follows. We use the pre-trained denoiser and the consistency model from ~\cite{edm} and ~\cite{song-consistency} respectively, with noise schedule $\sigma(t)=t$ and $t \in [\epsilon, T]$. % where $\epsilon=0.002, T=80$. 
Consider the corresponding PF ODE (from \cref{eqn:pfode}) for the above setting, as follows.
\begin{equation*}
    dx = - \frac{D_\theta(x,t)-x}{t} dt
\end{equation*}
%\begin{align*}\label{eqn:pfodect}
%    dx &= -\sigma(t) \dot{\sigma}(t) \nabla_{x_t}\log P(x_t) dt\\
%    &= -\sigma(t) \dot{\sigma}(t) \frac{D_\theta(x,t)-x}{\sigma^2(t)} dt\\
%    &= - \dot{\sigma}(t) \frac{D_\theta(x,t)-x}{\sigma(t)} dt\\
%    &= - \frac{D_\theta(x,t)-x}{t} dt
%    %x_0 &= C_{\theta}(x_t,t) =  x_t  + \int_{k=0}^{t} %\frac{D_\theta(x_k,k)-x_k}{k} dk\\
%\end{align*}
This PF ODE determines the trajectory and solving the trajectory origin $x_0$ involves solving the ODE above. Note that this $x_0$ is what the consistency model $C_\theta$ is trained to predict. As a rough approximation, solving the ODE with a backward Euler discretization step from $t$ to $0$, which essentially assumes the trajectory curves are linear, i.e. the Jacobian $\frac{dx}{dt}$ is constant for the interval $[0,t]$, and this gives $ C_\theta(x_t,t) = x_0 \approx D_\theta(x_t,t)$, with the approximation getting more accurate as the trajectory curve gets more linear. In an empirical setting, ~\cite{edm} also observes trajectories become more linear when $\sigma(t) \to 0$. While more analysis on this is still due, it motivates us to look at the denoiser as a proxy of the consistency model, which gradually becomes more and more accurate as $\sigma(t) \to 0$. Hence, we also consider settings that replace the consistency model with the denoiser in our algorithm.

\begin{algorithm}
\caption{MAP-GA for Image restoration}\label{alg:mapgainpaint}
%\SetKwData{Left}{left}\SetKwData{This}{this}\SetKwData{Up}{up}
%\SetKwFunction{Union}{Union}\SetKwFunction{FindCompress}{FindCompress}
\SetKwInOut{Input}{input}\SetKwInOut{Output}{output}

\Input{ time schedule: $\tau=[\ \tau_n, \tau_{n-1}, .. \tau_1, \tau_0\ ]$, \\noise schedule: $\sigma(.)$, denoiser: $D_{\theta},$\\
consistency model: $C_{\theta}$,\\
measurement: $y$, learning rate: $\lambda,$\\num gradient ascent iter: num\_iter,\\ boolean: use\_prior (default True),\\
%$HH^{\intercal}=\mathbb{I}_{m\times m}$
forward operator matrix: H,\\measurement noise: $\sigma_y$\\(Note: $\tau_0 = \epsilon $, $\tau_n = T $),
}
%{$\sigma=[\ \sigma_T,..\ \sigma_1,\sigma_0\ ]$}, \emph{$f_{\theta}$}, \emph{y}, \emph{num\_iter}, \emph{$\lambda$}}
{$z \sim \mathcal{N}(0,\sigma^2_{\tau_n}\mathbb{I})$}\\
\For{$i\ in\ (\ n, n-1,\ ..\ 1\ )$}
{
$t = \tau_i$\\
\For{$j\ in\ (\ 1, 2,\ ..\ num\_iter\ )$}{
$\hat{x}_\epsilon = C_{\theta}(z,t)$\\
%$grad_{likelihood} = \frac{H^{\intercal}(y-H\hat{x}_\epsilon)}{\sigma^2_y+ \sigma^2_{\epsilon}} $\\
$grad_{likelihood} = \frac{H^{\intercal}(\frac{\sigma^2_y}{\sigma^2_{\epsilon}}\mathbb{I}+HH^{\intercal})^{-1}(y-H\hat{x}_\epsilon)}{\sigma^2_{\epsilon}}$\\
\If{use\_prior}{$grad_{prior} = \frac{D_{\theta}(\hat{x}_\epsilon,\epsilon) - \hat{x}_\epsilon}{\sigma^2_{\epsilon}}$}
\Else{$grad_{prior}=0$}
$grad_{posterior} = grad_{likelihood} + grad_{prior} $\\
$grad = \left(\frac{\partial C_{\theta}(z,t)}{\partial z}\right)^{\intercal} grad_{posterior}$\\
$z = z + \lambda * grad$\\
}
$\hat{x}_\epsilon = C_{\theta}(z,t)$\\
$z = \mathcal{N}(\hat{x}_\epsilon, \sigma^2_{\tau_{i-1}} - \sigma^2_{\tau_{0}} \mathbb{I})$\\
}
\Output{z}
\end{algorithm}
When using \cref{alg:mapgainpaint} for noisy image restoration ($\sigma_y > 0$) in practice, we observe that it requires careful tuning of the learning rate and other hyperparameters. To avoid such additional and sensitive hyperparameters, we present \cref{alg:mapgainpaintnoisy} for noisy image restoration, based on our empirical observations in \cref{tab:table2}.
%The algorithm requires no additional hyperparameters and works well in practice. 
The motivation behind \cref{alg:mapgainpaintnoisy}, is to find an approximate solution using MAP-GA at diffusion time $t=\tau$ where $\sigma_\tau = \sigma_y$, and then use it as an initialization for PGDM~\cite{pgdm} for $\sigma_t < \sigma_y$. More specifically, we use MAP-GA until $\sigma_t = \sigma_y$, to estimate $x_{\sigma_y}$ and later use this as an initialization to PGDM for $\sigma_t < \sigma_y$. We do not use the prior term for the MAP-GA part in \cref{alg:mapgainpaintnoisy}, as we find it more effective.
%(Note that we could use $t$ and $\sigma_t$ interchangeably in our notations since our noise schedule is $\sigma(t)=t$). 
\begin{algorithm}
\caption{MAP-GA-PGDM Image restoration}\label{alg:mapgainpaintnoisy}
%\SetKwData{Left}{left}\SetKwData{This}{this}\SetKwData{Up}{up}
%\SetKwFunction{Union}{Union}\SetKwFunction{FindCompress}{FindCompress}
\SetKwInOut{Input}{input}\SetKwInOut{Output}{output}
\Input{ noise schedule: $\sigma(.)$, denoiser: $D_{\theta},$\\
consistency model: $C_{\theta}$,\\
measurement: $y$, learning rate: $\lambda,$\\num gradient ascent iter: num\_iter,\\%\\ boolean flag to use prior term: use\_prior,\\
forward operator matrix: H,\\measurement noise: $\sigma_y$\\
$\tau^{map-ga}=[\tau^{map}_n, \tau^{map}_{n-1}, .. \tau^{map}_1, \tau^{map}_0]$,\\
$\tau^{pgdm}=[\tau^{pgdm}_m, \tau^{pgdm}_{m-1}, .. \tau^{pgdm}_1, \tau^{pgdm}_0]$,\\
%(Note: $HH^{\intercal}=\mathbb{I}_{m\times m}$)\\
(Note: $\sigma_{\tau^{map}_0} = \sigma_y$ and $\tau^{map}_n=T$)\\
(Note: $\sigma_{\tau^{pgdm}_m} = \sigma_y$ and $\tau^{pgdm}_0 = \epsilon$),
}
%{$\sigma=[\ \sigma_T,..\ \sigma_1,\sigma_0\ ]$}, \emph{$f_{\theta}$}, \emph{y}, \emph{num\_iter}, \emph{$\lambda$}}
%$ $\\
$\text{ {\small=======MAP-GA=======}}$\\
{$z \sim \mathcal{N}(0,\sigma^2_{\tau^{map}_n}\mathbb{I})$}\\
\For{$i\ in\ (\ n, n-1,\ ..\ 1\ )$}
{
$t = \tau^{map}_i$\\
%\If{$\sigma_t > \sigma_y$}{
\For{$j\ in\ (\ 1, 2,\ ..\ num\_iter\ )$}{
$\hat{x}_\epsilon = C_{\theta}(z,t)$\\
$grad_{likelihood} = \frac{H^{\intercal}(\frac{\sigma^2_y}{\sigma^2_{\epsilon}}\mathbb{I}+HH^{\intercal})^{-1}(y-H\hat{x}_\epsilon)}{\sigma^2_{\epsilon}}$\\
%$grad_{likelihood} = \frac{H^{\intercal}(y-H\hat{x}_\epsilon)}{\sigma^2_y+ \sigma^2_{\epsilon}} $\\
%\If{use\_prior}{$grad_{prior} = \frac{D_{\theta}(\hat{x}_\epsilon,\epsilon) - \hat{x}_\epsilon}{\sigma^2_{\epsilon}}$}
%\Else{$grad_{prior}=0$}
%$grad_{prior}=0$\\
%$grad_{posterior} = grad_{likelihood} + grad_{prior}$\\
$grad_{posterior} = grad_{likelihood}$\\
$grad = \left(\frac{\partial C_{\theta}(z,t)}{\partial z}\right)^{\intercal} grad_{posterior}$\\
$z = z + \lambda * grad$\\
}

$\hat{x}_\epsilon = C_{\theta}(z,t)$\\
$z = \mathcal{N}(\hat{x}_\epsilon, \sigma^2_{\tau^{map}_{i-1}} - \sigma^2_{\tau^{map}_{0}} \mathbb{I})$\\
%}
%\Else{
%$x_t = z$\\
%use pgdm reverse step to get $x_{\tau_{i-1}}$\\
%$z = x_{\tau_{i-1}}$\\
%}
}
%$ $\\
$\text{ {\small========PGDM========}}$\\
$x_{\tau^{pgdm}_m} = z$\\
\For{$i\ in\ (\ m, m-1,\ ..\ 1\ )$}
{
$t = \tau^{pgdm}_i$\\
$\hat{x}_t = D_{\theta}(z,t)$\\
$\mu_t = \hat{x}_t + \sigma^2_t * \nabla_{x_t}\log P(y|x_t)$ \text{ 
   {\small*pgdm update*}}\\
$x_{\tau^{pgdm}_{i-1}} = \mathcal{N}(\mu_t, \sigma^2_{\tau^{pgdm}_{i-1}} - \sigma^2_{\tau^{pgdm}_{0}} \mathbb{I})$\\
}
\Output{$x_{\tau^{pgdm}_0}$}
\end{algorithm}
\begin{table*}[]
  \centering
  {\small{
  \resizebox{.65\textwidth}{!}{%
  \begin{tabular}{|lcccccccccccccccc|}
    \hline
     &  &  &  &  &  &  &  &  &  &  &  & & & & & \\
     & \multicolumn{2}{c}{box50} & \multicolumn{2}{c}{half} & \multicolumn{2}{c}{expand} & \multicolumn{2}{c}{box25} &  \multicolumn{2}{c}{sr2x} & \multicolumn{2}{c}{altlines} & \multicolumn{2}{c}{deblur} & \multicolumn{2}{c|}{supres4x} \\
     
    \ \ \ \ Method & FID$\downarrow$ & LPIPS$\downarrow$ & FID$\downarrow$ & LPIPS$\downarrow$ & FID$\downarrow$ & LPIPS$\downarrow$ & FID$\downarrow$ & LPIPS$\downarrow$ & FID$\downarrow$ & LPIPS$\downarrow$ & FID$\downarrow$ & LPIPS$\downarrow$ & FID$\downarrow$ & LPIPS$\downarrow$ & FID$\downarrow$ & LPIPS$\downarrow$\\
    &  &  &  &  &  &  &  &  &  &  &  &  & & & & \\\hdashline
    &  &  &  &  &  &  &  &  &  &  &  &  & & & & \\
    %&  &  &  &  &  &  &  &  &  &  &  &  \\
    MAP-GA(D,NP) & $\textbf{32.312}$ & $\textbf{0.096}$ & $\textbf{37.986}$ & $\textbf{0.157}$ & $93.068$ & $0.419$ & $\textbf{11.019}$ & $\textbf{0.019}$ & $\textbf{30.072}$ & $\textbf{0.034}$ & $\textbf{18.282}$ & $\textbf{0.015}$ & $\textbf{19.704}$ & $\textbf{0.007}$ & $66.968$ & $0.148$\\
    &  &  &  &  &  &  &  &  &  &  &  & &  &  &  & \\
    MAP-GA(D) & $35.761$ & $0.100$ & $43.315$ & $0.164$ & $106.32$ & $0.430$ & $11.169$ & $0.019$ & $33.188$ & $0.036$ & $19.103$ & $0.016$ & $19.779$ & $0.008$ & $78.791$ & $0.179$\\
    &  &  &  &  &  &  &  &  &  &  &  &  &  &  &  & \\
    MAP-GA(NP) & $34.944$ & $0.113$ & $39.243$ & $0.162$ & $\textbf{69.004}$ & $\textbf{0.388}$ & $12.818$ & $0.027$ & $30.303$ & $0.035$ & $20.321$ & $0.018$ & $22.090$ & $0.008$ & $\textbf{46.349}$ & $\textbf{0.112}$\\
    &  &  &  &  &  &  &  &  &  &  &  &  &  &  &  & \\
    MAP-GA & $36.733$ & $0.113$ & $41.151$ & $0.164$ & $87.952$ & $0.400$ & $12.836$ & $0.027$ & $33.752$ & $0.036$ & $20.895$ & $0.018$ & $21.314$ & $0.010$ & $59.624$ & $0.120$\\
    &  &  &  &  &  &  &  &  &  &  &  &  &  &  &  & \\
    PGDM~\cite{pgdm} & $49.370$ & $0.151$ & $54.261$ & $0.245$ & $127.95$ & $0.479$ & $14.255$ & $0.021$ & $38.433$ & $0.046$ & $20.446$ & $0.019$ & $19.857$ & $0.007$ & $89.614$ & $0.238$\\
    &  &  &  &  &  &  &  &  &  &  &  &  &  &  &  & \\
    DDRM~\cite{ddrm} & $51.477$ & $0.165$ & $56.643$ & $0.264$ & $136.06$ & $0.492$ & $15.000$ & $0.023$ & $35.033$ & $0.041$ & $19.331$ & $0.017$ & $23.195$ & $0.009$ & $78.712$ & $0.235$\\
    &  &  &  &  &  &  &  &  &  &  &  &  &  &  &  & \\
    CT-ZSIE~\cite{song-consistency} & $38.017$ & $0.129$ & $44.152$ & $0.191$ & $70.634$ & $0.424$ & $13.040$ & $0.025$ & $42.500$ & $0.060$ & $26.737$ & $0.029$ & $29.223$ & $0.013$ & $56.698$ & $0.116$\\
    &  &  &  &  &  &  &  &  &  &  &  &  &  &  &  & \\
    \hline
  \end{tabular}%
  }
  }}
  \caption{ Noiseless image restoration on ImageNet64 1K validation set using {\footnotesize MAP-GA} and variants. The setting {\footnotesize MAP-GA} denotes the default \cref{alg:mapgainpaint}, {\footnotesize MAP-GA(NP)} denotes {\footnotesize MAP-GA} with no prior, {\footnotesize MAP-GA(D)} denotes {\footnotesize MAP-GA} with denoiser replacing the consistency model, and {\footnotesize MAP-GA(D,NP)} denotes {\footnotesize MAP-GA} with no prior and the denoiser replacing the consistency model. }
  \label{tab:table1}
\end{table*}
\begin{table*}[]
  \centering
  {\small{
  \resizebox{.65\textwidth}{!}{%
  \begin{tabular}{|lcccccccccccccccc|}
    \hline
     &  &  &  &  &  &  &  &  &  &  &  &  &  &  &  & \\
     & \multicolumn{2}{c}{box50} & \multicolumn{2}{c}{half} & \multicolumn{2}{c}{expand} & \multicolumn{2}{c}{box25} &  \multicolumn{2}{c}{sr2x} & \multicolumn{2}{c}{altlines} & \multicolumn{2}{c}{deblur} & \multicolumn{2}{c|}{supres4x} \\
     
    \ \ \ \ Method & FID$\downarrow$ & LPIPS$\downarrow$ & FID$\downarrow$ & LPIPS$\downarrow$ & FID$\downarrow$ & LPIPS$\downarrow$ & FID$\downarrow$ & LPIPS$\downarrow$ & FID$\downarrow$ & LPIPS$\downarrow$ & FID$\downarrow$ & LPIPS$\downarrow$ & FID$\downarrow$ & LPIPS$\downarrow$ & FID$\downarrow$ & LPIPS$\downarrow$ \\
    &  &  &  &  &  &  &  &  &  &  &  &  &  &  &  &  \\\hdashline
    &  &  &  &  &  &  &  &  &  &  &  &  &  &  &  & \\
    %&  &  &  &  &  &  &  &  &  &  &  &  \\
    MAP-GA(D,NP) & $\textbf{70.880}$ & $\textbf{0.128}$ & $\textbf{68.022}$ & $\textbf{0.269}$ & $174.96$ & $0.695$ & $18.221$ & $0.028$ & $\textbf{22.819}$ & $0.059$ & $13.632$ & $0.036$ & $81.461$ & $0.214$ & $59.895$ & $0.205$ \\
    &  &  &  &  &  &  &  &  &  &  &  &  &  & &  & \\
    MAP-GA(D) & $74.995$ & $0.137$ & $75.748$ & $0.280$ & $197.91$ & $0.696$ & $\textbf{16.795}$ & $\textbf{0.025}$ & $24.420$ & $0.063$ & $13.669$ & $0.033$ & $86.585$ & $0.220$ & $63.308$ & $0.214$ \\
    &  &  &  &  &  &  &  &  &  &  &  &  &  & &  & \\
    MAP-GA(NP) & $98.749$ & $0.165$ & $86.975$ & $0.297$ & $\textbf{155.55}$ & $\textbf{0.645}$ & $30.788$ & $0.046$ & $24.434$ & $\textbf{0.054}$ & $16.126$ & $0.034$ & $\textbf{80.337}$ & $\textbf{0.211}$ & $\textbf{49.622}$ & $\textbf{0.146}$ \\
    &  &  &  &  &  &  &  &  &  &  &  &  &  & &  & \\
    MAP-GA & $108.87$ & $0.171$ & $85.936$ & $0.291$ & $175.27$ & $0.652$ & $34.711$ & $0.053$ & $25.666$ & $0.056$ & $15.139$ & $0.036$ & $84.971$ & $0.214$ & $68.156$ & $0.169$ \\
    &  &  &  &  &  &  &  &  &  &  &  &  &  & &  & \\
    PGDM~\cite{pgdm} & $120.82$ & $0.194$ & $94.920$ & $0.360$ & $227.97$ & $0.765$ & $28.357$ & $0.041$ & $27.927$ & $0.070$ & $14.211$ & $0.037$ & $94.629$ & $0.227$ & $77.533$ & $0.248$ \\
    &  &  &  &  &  &  &  &  &  &  &  &  &  & &  & \\
    DDRM~\cite{ddrm} & $131.86$ & $0.198$ & $101.86$ & $0.379$ & $224.93$ & $0.778$ & $29.571$ & $0.041$ & $23.686$ & $0.065$ & $\textbf{12.040}$ & $\textbf{0.026}$ & $105.28$ & $0.266$ & $75.923$ & $0.251$ \\
    &  &  &  &  &  &  &  &  &  &  &  &  &  & &  & \\
    CT-ZSIE~\cite{song-consistency} & $118.27$ & $0.209$ & $121.11$ & $0.375$ & $200.03$ & $0.704$ & $34.246$ & $0.046$ & $47.353$ & $0.145$ & $24.663$ & $0.070$ & $97.343$ & $0.264$ & $50.608$ & $0.157$ \\
    &  &  &  &  &  &  &  &  &  &  &  &  &  & &  & \\
    \hline
  \end{tabular}%
  }
  }}
  \caption{ Noiseless image restoration on $100$ LSUNCat256 images using {\footnotesize MAP-GA} and variants. The setting {\footnotesize MAP-GA} denotes the default \cref{alg:mapgainpaint}, {\footnotesize MAP-GA(NP)} denotes {\footnotesize MAP-GA} with no prior, {\footnotesize MAP-GA(D)} denotes {\footnotesize MAP-GA} with denoiser replacing the consistency model, and {\footnotesize MAP-GA(D,NP)} denotes {\footnotesize MAP-GA} with no prior and the denoiser replacing the consistency model. }
  \label{tab:table2:lsuncat}
\end{table*}
\begin{figure*}[h!]
    \centering
    \resizebox{.65\textwidth}{!}{%
    \includegraphics[keepaspectratio]{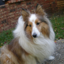}
    \includegraphics[keepaspectratio]{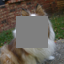}
    \includegraphics[keepaspectratio]{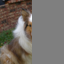}
    \includegraphics[keepaspectratio]{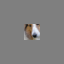}
    \includegraphics[keepaspectratio]{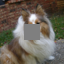}
    \includegraphics[keepaspectratio]{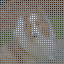}
    \includegraphics[keepaspectratio]{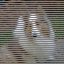}%
    }
    \caption{Left to right: original image, and mask settings: box50, half, expand, box25, sr2x, altlines}
    \label{fig:masksettings}
\end{figure*}
\section{Experiments}\label{sec:experiments}
We consider the tasks of image inpainting, deblurring, and $4\times$ super-resolution. Inspired by ~\cite{repaint}, for image inpainting, we evaluate the performance across six different mask settings (\cref{fig:masksettings}) denoting varying levels of degradation. The mask settings {\it box50} and {\it box25} indicate a square crop at the center of the image, with the crop width equal to {\it 50\%} and {\it 25\%} of the image width respectively. In {\it half}, we mask out the right half of the image, {\it expand} is the complement of {\it box25}. {\it sr2x} denotes a $2\times$ super-resolution mask, and {\it altlines} masks out alternate rows of pixels. We evaluate the restored images using FID and LPIPS metrics.\\

We evaluate the performance of MAP-GA (\cref{alg:mapgainpaint}) on ImageNet~\cite{imagenet} 1K validation set with $64\times 64$ resolution (in \cref{tab:table1}) and on $100$ random images of LSUNCat~\cite{lsun} with $256\times 256$ resolution (in \cref{tab:table2:lsuncat}). We use the pre-trained denoisers and the consistency models from ~\cite{edm,song-consistency} with their default settings. In the experiments, the setting MAP-GA denotes the default \cref{alg:mapgainpaint}, MAP-GA(NP) denotes MAP-GA with no prior, MAP-GA(D) denotes MAP-GA with denoiser replacing the consistency model, and MAP-GA(D,NP) denotes MAP-GA with no prior and the denoiser replacing the consistency model. We compare against DDRM~\cite{ddrm} and PGDM~\cite{pgdm} (both only use the denoiser) and against the zero-shot image editing (CT-ZSIE) algorithm proposed in ~\cite{song-consistency} which only use the consistency model. The results from \cref{tab:table1,tab:table2:lsuncat} show MAP-GA and variants outperform DDRM, PGDM, and CT-ZSIE with a significant margin on several tasks.\\

Note that MAP-GA uses gradient ascent (first-order gradient-based optimization algorithm) to optimize the underlying MAP objective which is typically highly non-convex and is not guaranteed to find the global optima. However, it could converge with an initialization closer to the global optima. To corroborate this, we design the following toy experiment. We consider the noiseless inpainting task from earlier, but, in \cref{alg:mapgainpaint}, we set $\tau_n = \hat{\tau} \ll T$, instead of the default setting $\tau_n=T$ and we also initialize $z \sim \mathcal{N}(x_0,\sigma^2_{\hat{\tau}}\mathbb{I})$, where $x_0$ is the corresponding ground truth image for the measurement $y$. From \cref{tab:table2}, we observe significant improvements in the performance. (We use $\hat{\tau}=0.5$, note that $\sigma_{0.5}=0.5$, as we use the noise schedule $\sigma_t=t$). This shows that MAP-GA is only limited by the choice of optimizer and reinforces the need for better optimization algorithms. While it is important to consider better design choices (for the schedules and hyperparameters), adaptive-gradient-based optimizers (such as momentum, Adam), or higher-order methods, we leave this for future work as it requires a thorough analysis.\\

In \cref{tab:table3}, we report the results on noisy inpainting using \cref{alg:mapgainpaintnoisy} on the ImageNet64 1K validation set. In \cref{alg:mapgainpaintnoisy}, and all its variants, we fix $m=20$ time steps for PGDM. Even with high levels of measurement noise, MAP-GA-PGDM shows promising improvements over PGDM. In all our experiments, for MAP-GA and variants, we fix a budget of 1000 steps and run ablations for $(num\_steps, num\_iter)$ from the set \{(20,50),(50,20),(100,10),(200,5),(250,4),(500,2),(1000,1)\}. For DDRM, PGDM, and CT-ZSIE, we run ablations for $num\_steps$ from the set \{20,50,100,200,250,500,1000\}. The learning rate in all our experiments was set to $\sigma^2_y + \sigma^2_\epsilon$.
%We follow a time-step schedule similar to ~\cite{edm}, with their default settings. The best results for all our methods resulted from the ablation (20,50) denoting 20 time-steps with 50 gradient ascent iterations. For PGDM~\cite{pgdm}, each setting from our results table is , and most of the runs have their peak performance at $500$ steps. The performance reported in the tables is the best result for each method over the ablations for (num\_time\_steps, num\_iter). .
%$\sigma_y$ indicates the measurement noise and the setting MAP-GA-PGDM denote the default \cref{alg:mapgainpaintnoisy}, MAP-GA-PGDM(D) denote MAP-GA-PGDM with denoiser replacing the consistency model. 

\begin{table*}[]
  \centering
  {\small{
  \resizebox{.65\textwidth}{!}{%
  \begin{tabular}{|lcccccccccccc|}
    \hline
     &  &  &  &  &  &  &  &  &  &  &  &  \\
     & \multicolumn{2}{c}{box50} & \multicolumn{2}{c}{half} & \multicolumn{2}{c}{expand} & \multicolumn{2}{c}{box25} &  \multicolumn{2}{c}{sr2x} & \multicolumn{2}{c|}{altlines} \\
     
    \ \ \ \ Method & FID$\downarrow$ & LPIPS$\downarrow$ & FID$\downarrow$ & LPIPS$\downarrow$ & FID$\downarrow$ & LPIPS$\downarrow$ & FID$\downarrow$ & LPIPS$\downarrow$ & FID$\downarrow$ & LPIPS$\downarrow$ & FID$\downarrow$ & LPIPS$\downarrow$\\
    &  &  &  &  &  &  &  &  &  &  &  &  \\\hdashline
    &  &  &  &  &  &  &  &  &  &  &  &  \\
    %&  &  &  &  &  &  &  &  &  &  &  &  \\
    MAP-GA(D,NP) & $26.661$ & $0.043$ & $30.621$ & $0.064$ & $73.510$ & $0.155$ & $\textbf{9.165}$ & $0.010$ & $29.878$ & $0.027$ & $\textbf{16.971}$ & $\textbf{0.012}$ \\
    &  &  &  &  &  &  &  &  &  &  &  &  \\
    MAP-GA(D) & $34.689$ & $0.052$ & $41.098$ & $0.083$ & $90.768$ & $0.175$ & $10.891$ & $0.011$ & $34.334$ & $0.031$ & $18.718$ & $0.013$ \\
    &  &  &  &  &  &  &  &  &  &  &  &  \\
    MAP-GA(NP) & $\textbf{25.205}$ & $\textbf{0.043}$ & $\textbf{27.208}$ & $\textbf{0.047}$ & $\textbf{42.671}$ & $\textbf{0.075}$ & $10.936$ & $0.017$ & $\textbf{27.496}$ & $\textbf{0.023}$ & $18.732$ & $0.013$ \\
    &  &  &  &  &  &  &  &  &  &  &  &  \\
    MAP-GA & $29.508$ & $0.045$ & $33.653$ & $0.050$ & $73.318$ & $0.106$ & $11.257$ & $0.016$ & $34.364$ & $0.026$ & $20.041$ & $0.013$ \\
    &  &  &  &  &  &  &  &  &  &  &  &  \\
    PGDM~\cite{pgdm} &  $31.952$ & $0.056$ & $36.639$ & $0.082$ & $71.352$ & $0.138$ & $10.886$ & $\textbf{0.009}$ & $33.199$ & $0.041$ & $19.551$ & $0.018$ \\
    &  &  &  &  &  &  &  &  &  &  &  &  \\
    \hline
  \end{tabular}%
  }
  }}
  \caption{ Noiseless inpainting on ImageNet64 1K validation set. Using the ground truth image ($x_0$) for the measurement $y$, we create a sample at $t=0.5$ via ($x_{0.5} = x_0 + 0.5 * \eta$, where, $\eta \sim \mathcal{N}(0,\mathbb{I}))$ and initialize \cref{alg:mapgainpaint} with $z=x_{0.5}$, and $\tau_n=0.5$ }
  \label{tab:table2}
\end{table*}

 \begin{table*}[]
  \centering
  {\small{
  \resizebox{.65\textwidth}{!}{%
  \begin{tabular}{|lccccccccccccc|}
    \hline
     &  &  &  &  &  &  &  &  &  &  &  &  &  \\
     & & \multicolumn{2}{c}{box50} & \multicolumn{2}{c}{half} & \multicolumn{2}{c}{expand} & \multicolumn{2}{c}{box25} &  \multicolumn{2}{c}{sr2x} & \multicolumn{2}{c|}{altlines} \\
    \ \ \ \ Method & $\sigma_y$ & FID$\downarrow$ & LPIPS$\downarrow$ & FID$\downarrow$ & LPIPS$\downarrow$ & FID$\downarrow$ & LPIPS$\downarrow$ & FID$\downarrow$ & LPIPS$\downarrow$ & FID$\downarrow$ & LPIPS$\downarrow$ & FID$\downarrow$ & LPIPS$\downarrow$\\
    &  &  &  &  &  &  & &  &  &  &  &  &   \\\hdashline
    &  &  &  &  &  &  & &  &  &  &  &  &   \\
    %&  &  &  &  &  &  &  &  &  &  &  &  \\
    MAP-GA-PGDM(D) & $0.05$ & $\textbf{56.173}$ & $\textbf{0.125}$ & $62.026$ & $0.193$ & $108.16$ & $0.438$ & $\textbf{38.725}$ & $\textbf{0.039}$ & $57.41$ & $0.080$ & $46.046$ & $0.046$\\
    &  &  &  &  &  &  &  &  &  &  &  & &  \\
    MAP-GA-PGDM & $0.05$ & $58.283$ & $0.147$ & $\textbf{61.588}$ & $\textbf{0.184}$ & $\textbf{91.508}$ & $\textbf{0.406}$ & $44.667$ & $0.085$ & $\textbf{57.308}$ & $\textbf{0.073}$ & $\textbf{45.265}$ & $\textbf{0.042}$\\
    &  &  &  &  &  &  &  &  &  &  &  & &  \\
    PGDM~\cite{pgdm} & $0.05$ & $77.824$ & $0.175$ & $80.248$ & $0.257$ & $135.99$ & $0.495$ & $53.136$ & $0.049$ & $86.289$ & $0.126$ & $66.203$ & $0.066$\\
    &  &  &  &  &  &  &  &  &  &  &  & &  \\\hdashline
    &  &  &  &  &  &  &  &  &  &  &  & &  \\
    
    MAP-GA-PGDM(D) & $0.1$ & $\textbf{72.543}$ & $\textbf{0.166}$ & $79.535$ & $0.230$ & $114.51$ & $0.464$ & $\textbf{57.556}$ & $\textbf{0.076}$ & $76.733$ & $0.145$ & $65.154$ & $0.096$\\
    &  &  &  &  &  &  &  &  &  &  &  & &  \\
    MAP-GA-PGDM & $0.1$  & $74.130$ & $0.191$ & $\textbf{78.322}$ & $\textbf{0.225}$ & $\textbf{103.85}$ & $\textbf{0.440}$ & $65.720$ & $0.161$ & $\textbf{76.248}$ & $\textbf{0.134}$ & $\textbf{63.134}$ & $\textbf{0.089}$\\
    &  &  &  &  &  &  &  &  &  &  &  & &  \\
    PGDM~\cite{pgdm} & $0.1$ & $96.485$ & $0.216$ & $99.170$ & $0.286$ & $145.40$ & $0.519$ & $78.620$ & $0.100$ & $109.47$ & $0.231$ & $90.925$ & $0.138$\\
    &  &  &  &  &  &  &  &  &  &  &  & &  \\\hdashline
    %&  &  &  &  &  &  &  &  &  &  &  &  &  \\
    
    %MAP-GA-PGDM(D) & $0.2$ & $91.112$ & $\textbf{0.243}$ & $99.432$ & $0.297$ & $119.45$ & $0.496$ & $\textbf{79.329}$ & $\textbf{0.154}$ & $110.48$ & $0.260$ & $87.727$ & $0.188$\\
    %&  &  &  &  &  & &  &  &  &  &  &  &  \\
    %MAP-GA-PGDM & $0.2$ & $\textbf{91.087}$ & $0.252$ & $\textbf{97.531}$ & $\textbf{0.296}$ & $\textbf{116.78}$ & $\textbf{0.491}$ & $80.647$ & $0.177$ & $\textbf{103.66}$ & $\textbf{0.248}$ & $\textbf{86.174}$ & $\textbf{0.186}$\\
    %&  &  &  &  & &  &  &  &  &  &  &  &  \\
    %PGDM~\cite{pgdm} & $0.2$ & $116.81$ & $0.292$ & $117.03$ & $0.344$ & $159.17$ & $0.562$ & $103.43$ & $0.193$ & $132.89$ & $0.366$ & $114.54$ & $0.255$\\
    %&  &  &  & &  &  &  &  &  &  &  &  &  \\
    \hline
  \end{tabular}%
  }
  }}
  \caption{ Noisy inpainting on ImageNet64 1K validation set. $\sigma_y$ denotes the measurement noise. The setting {\footnotesize MAP-GA-PGDM} denotes the default \cref{alg:mapgainpaintnoisy}, {\footnotesize MAP-GA-PGDM(D)} denote {\footnotesize MAP-GA-PGDM} with denoiser replacing the consistency model. }
  \label{tab:table3}
\end{table*}

% Also now, these results reveal that replacing the consistency model with the Denoiser performs worse in most settings, contrary to the observations from \cref{tab:table1}. This reinforces the need for better optimization algorithms to draw plausible conclusions from ablations. We also observe that using \cref{alg:mapgainpaint} for Noisy Image Inpainting (i.e. for $\sigma_y > 0$) requires heavy tuning of the learning rate and time-step schedules.\\
%Our ablations for replacing the consistency model with the Denoiser, on the contrary, show improved performance in most settings, except in the case of {\it expand} mask, where the performance is better when using a consistency model. Also, our ablations with/without the prior term show improved performance when not using the "prior" term. However, we refrain from drawing far-fetched conclusions regarding the ablations because of the following. Note that we are optimizing the underlying MAP objective (typically highly non-convex) using gradient ascent (first-order gradient-based optimization algorithm) which in practice needs careful tuning of the learning rate, time-step schedules, initialization, and other hyperparameters.\\ 

\section{Discussion}\label{sec:discussion}
\subsection{Runtime comparison}
\cref{tab:runtime} compares the wall-clock time of MAP-GA and variants against DDRM, PGDM, and CT-ZSIE for image inpainting on ImageNet64. To ensure a fair comparison, we keep the batch size fixed at 50 for all the methods and compare their runtime per iteration. For DDRM, CT-ZSIE, and PGDM, it is the total runtime divided by $num\_steps$ (i.e. reverse diffusion time steps), while for MAP-GA and variants, it is the effective runtime per $num\_steps$ per $num\_iter$ (i.e. it is the runtime when $num\_steps = num\_iter = 1$). MAP-GA and variants are $1.5\times$ to $2\times$ slower per iteration than PGDM and $3\times$ to $4\times$ slower than DDRM and CT-ZSIE.
\begin{table}[h]
    \centering
    {\small{\resizebox{.30\textwidth}{!}{\begin{tabular}{|c|c|}%
    \hline
        Method  & Runtime per iteration \\
        \hline
        DDRM~\cite{ddrm} & 150 ms\\
        \hline 
        CT-ZSIE~\cite{song-consistency} & 150 ms \\
        \hline
        PGDM~\cite{pgdm} & 304 ms\\
        \hline
        MAP-GA(D,NP) & 456 ms\\
        \hline
        MAP-GA(D) & 602 ms\\
        \hline
        MAP-GA(NP) & 455 ms\\
        \hline
        MAP-GA & 603 ms\\
        \hline
    \end{tabular}%
    }}
    }
    \caption{Runtime comparison on NVIDIA A100 40GB GPU}
    \label{tab:runtime}
\end{table}
\subsection{Concurrent works}
The algorithms presented in this paper are similar to ZSIR~\cite{zsir}, and DMPlug~\cite{dmplug} from a practical perspective. However, our MAP formulation is novel, has a strong theoretical motivation, and connects the PF ODE and the consistency model with the MAP optimization for solving inverse problems. Unlike ZSIR and DMPlug, we consider the prior term and show that the gradient of the log-prior is tractable, making the gradient of the log-posterior tractable. We show that MAP-GA is only limited by the optimizer's choice in practice. ZSIR and DMPlug approximate the PF ODE trajectory origin $x_0$ with a multi-step approximation using the denoiser and hence require backpropagation through the chain of cascaded functions to optimize for the loss objective. In our case, all MAP-GA variants require a single vector-Jacobian product per iteration. 
%and our proposed framework has several directions for future research, such as improving optimization with Hessian-based methods, more analysis on local optima, approximating the consistency model with the denoiser among others. Also, our method and algorithms are valid for general SDE-based Diffusion models, and for more general inverse problems. %than the inpainting task we considered.%While we compare only against PGDM~\cite{pgdm}, it is the most competitive baseline and can also be seen from the experiments in DMPlug~\cite{dmplug}. In this work, we analyze our method extensively, on image inpainting with various degradation masks. However, the experiments from DMPlug~\cite{dmplug} further complement and validate our work. 

%While our MAP formulation ideally requires both the consistency model and the Denoiser, we also have presented a strong motivation to consider the Denoiser as a proxy of the consistency model. 
 %The reason we do not consider the prior in some of our algorithms is purely due to the challenging optimization in practice, which could be mitigated with better optimization strategies, which is something we did not explore in this work.Moreover, 

\begin{figure}[]
    \centering
    %\resizebox{.5\linewidth}{.25\textwidth}{%
    \includegraphics[width=0.20\linewidth,keepaspectratio]{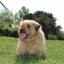}
    \includegraphics[width=0.20\linewidth,keepaspectratio]{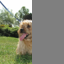}
    \includegraphics[width=0.20\linewidth,keepaspectratio]{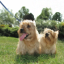}
    \includegraphics[width=0.20\linewidth,keepaspectratio]{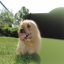}
    \includegraphics[width=0.20\linewidth,keepaspectratio]{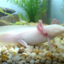}
    \includegraphics[width=0.20\linewidth,keepaspectratio]{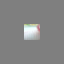}
    \includegraphics[width=0.20\linewidth,keepaspectratio]{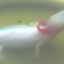}
    \includegraphics[width=0.20\linewidth,keepaspectratio]{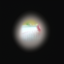}
    \includegraphics[width=0.20\linewidth,keepaspectratio]{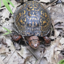}
    \includegraphics[width=0.20\linewidth,keepaspectratio]{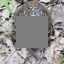}
    \includegraphics[width=0.20\linewidth,keepaspectratio]{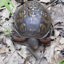}
    \includegraphics[width=0.20\linewidth,keepaspectratio]{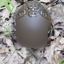}%
    %}
    \caption{Noiseless inpainting task. Left to right: original image, masked image, MAP-GA(\cref{alg:mapgainpaint}), PGDM($\sigma_y=0$). Top to bottom: \emph{half mask}, \emph{expand mask}, \emph{box50 mask}}
    \label{fig:enter-label}
\end{figure}%

\begin{figure}[]
    \centering
    \includegraphics[width=0.20\linewidth,keepaspectratio]{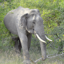}
    \includegraphics[width=0.20\linewidth,keepaspectratio]{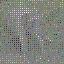}
    \includegraphics[width=0.20\linewidth,keepaspectratio]{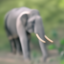}
    \includegraphics[width=0.20\linewidth,keepaspectratio]{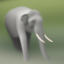}
    \includegraphics[width=0.20\linewidth,keepaspectratio]{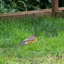}
    \includegraphics[width=0.20\linewidth,keepaspectratio]{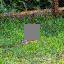}
    \includegraphics[width=0.20\linewidth,keepaspectratio]{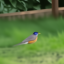}
    \includegraphics[width=0.20\linewidth,keepaspectratio]{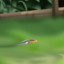}
    \caption{Noisy inpainting task. Left to right: original image, masked image, MAP-GA-PGDM(\cref{alg:mapgainpaintnoisy}, $\sigma_y=0.1$), PGDM($\sigma_y=0.1$). Top to bottom: \emph{sr2x mask}, \emph{box25 mask}}
    \label{fig:enter-label}
\end{figure}

\section{Conclusion}\label{sec:conclusion}

In this paper, we proposed a novel MAP formulation for solving inverse problems using pre-trained unconditional diffusion models. Note that conditional generation is a core requirement in solving inverse problems. We connect the Probability Flow ODE and the consistency model with the optimization process for the MAP objective in tasks that involve conditional generation. We show that the gradient of the MAP objective is tractable, allowing the use of gradient-based optimization methods. To use our framework in practice, we propose an algorithm with a multi-step gradient ascent strategy for MAP optimization. We propose practically effective algorithms for image restoration using our framework (extendable to other general inverse problems). We validate our algorithms with extensive experiments on image deblurring, super-resolution, and inpainting.%Since we use gradient-ascent (a simple first-order gradient method), we observed that extensive tuning of the hyperparameters is required due to the challenging optimization objective.  

%\begin{comment}
\section*{Acknowledgements}
This work is part of the Marie Skłodowska-Curie Actions project \emph{MODELAIR}, funded by the European Commission under the Horizon Europe program through grant agreement no. 101072559. The computations and the data handling were enabled by resources provided by the National Academic Infrastructure for Supercomputing in Sweden (NAISS), partially funded by the Swedish Research Council through grant agreement no. 2022-06725. Bharath thanks Sebastian Gerard and Heng Fang for their feedback on improving the presentation of this paper.
{\small
\bibliographystyle{ieee_fullname}
\bibliography{egbib}
}
\end{document}

\title{Appendix for Inverse Problems with Diffusion Models: A MAP Estimation Perspective}  % **** Enter the paper title here
\maketitle
\appendix

\section{Runtime Analysis}
In MAP-GA (Algorithm 2), per iteration i.e. when $num\_steps = num\_iter = 1$, we have 2 NFE (Neural Function Evaluations i.e. forward pass for inference) of the consistency model, 1 NFE of the denoiser and a vector-Jacobian product computation using the consistency model. Computational cost is 2*$O(C_\theta) + O(D_\theta) + O(vjpC_\theta)$, where $O(C_\theta)$ (for a lack of better notation), denotes the cost of 1 NFE of the consistency model, $O(D_\theta)$ denotes the cost of 1 NFE of the denoiser and $O(vjpC_\theta)$ denotes the cost of 1 backward pass for the vector-Jacobian product using the consistency model.\\

Similarly, the computational costs per iteration for MAP-GA variants, PGDM, DDRM, and CT-ZSIE are reported in Tab 1. (We assume that in MAP-GA and variants, and PGDM, the cost for computing the terms such as $grad_{likelihood}$ involving the matrix multiplication and inverse is negligible compared to an NFE. We use efficient SVD decomposition of the forward operator matrix $H$ for several image restoration problems to make matrix inverse computation efficient. So we assume that terms like $grad_{likelihood}$ can be computed efficiently in practice).\\

In our case, the pre-trained denoiser and the consistency models have an exactly similar model architecture (except in the final layer), so the computational costs for using the denoiser and the consistency model are roughly similar i.e. $O(C_\theta) \approx O(D_\theta)$, and $O(vjpC_\theta) \approx O(vjpD_\theta)$. The runtime comparison from Tab 5 in the main paper also validates the table below.\\

\begin{table}[]
    \centering
    \begin{tabular}{|c|c|}
    \hline
       Method & Computational cost per iteration \\\hline
       MAP-GA  &  $2*O(C_\theta) + O(D_\theta) + O(vjpC_\theta)$ \\\hline
       MAP-GA(D)  & $2*O(D_\theta) + O(D_\theta) + O(vjpD_\theta)$ \\\hline
       MAP-GA(NP) & $2*O(C_\theta) + O(vjpC_\theta)$\\\hline
       MAP-GA(D,NP) & $2*O(D_\theta) + O(vjpD_\theta)$\\\hline
       PGDM & $O(D_\theta) + O(vjpD_\theta)$\\\hline
       DDRM & $O(D_\theta)$\\\hline
       CT-ZSIE & $O(C_\theta)$\\\hline
    \end{tabular}
    \caption{Computational cost of MAP-GA ( and variants), and other baselines. Note that MAP-GA(D) and MAP-GA(D,NP) indicate that the consistency model is replaced with the denoiser in MAP-GA and MAP-GA(NP) respectively. One iteration in PGDM, DDRM, and CT-ZSIE means one backward diffusion step.}
    \label{tab:my_label}
\end{table}
\newpage
\section{Qualitative examples}
\begin{figure*}[h]
    \centering
    %\resizebox{.5\linewidth}{.25\textwidth}{%
    \includegraphics[width=0.30\textwidth,keepaspectratio]{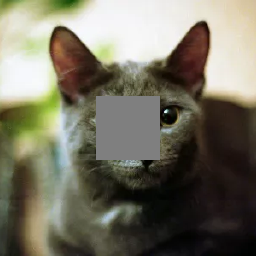}
    \includegraphics[width=0.30\textwidth,keepaspectratio]{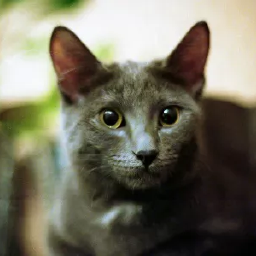}
    \includegraphics[width=0.30\textwidth,keepaspectratio]{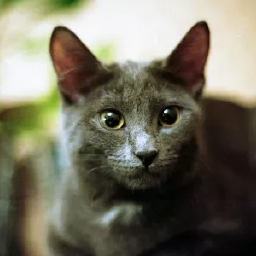}
    \includegraphics[width=0.30\textwidth,keepaspectratio]{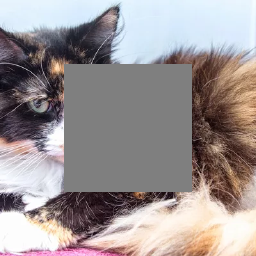}
    \includegraphics[width=0.30\textwidth,keepaspectratio]{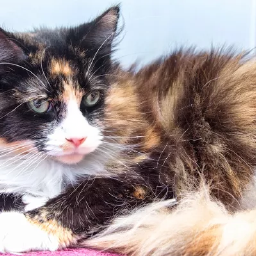}
    \includegraphics[width=0.30\textwidth,keepaspectratio]{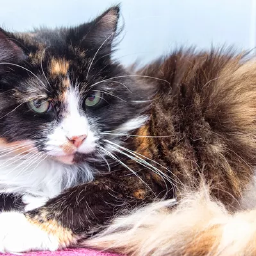}
    \includegraphics[width=0.30\textwidth,keepaspectratio]{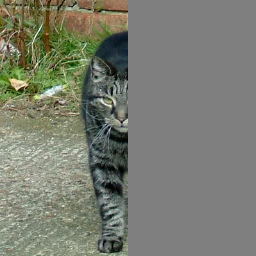}
    \includegraphics[width=0.30\textwidth,keepaspectratio]{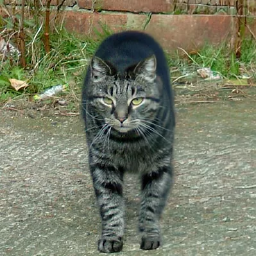}
    \includegraphics[width=0.30\textwidth,keepaspectratio]{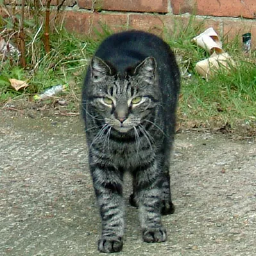}

    \caption{ MAP-GA(D,NP) based image restoration on LSUNCat256. Left to right: observed, recovered, original images. Top to bottom: mask settings {\it box25}, {\it box50}, {\it half}. We choose $num\_steps=20$, $num\_iter=50$}
    \label{fig:enter-label}
\end{figure*}%

\begin{figure*}
    \centering
    %\resizebox{.5\linewidth}{.25\textwidth}{%
    \includegraphics[width=0.30\textwidth,keepaspectratio]{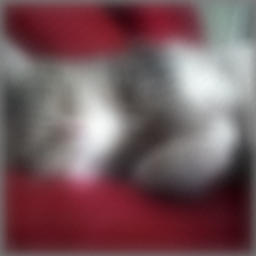}
    \includegraphics[width=0.30\textwidth,keepaspectratio]{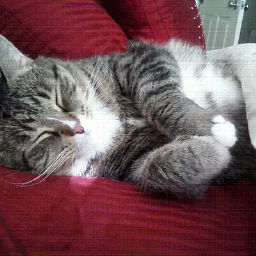}
    \includegraphics[width=0.30\textwidth,keepaspectratio]{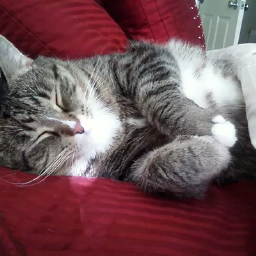}

    \includegraphics[width=0.30\textwidth,keepaspectratio]{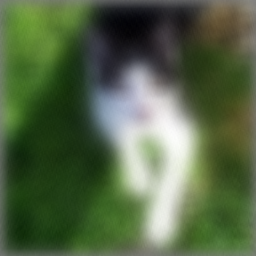}
    \includegraphics[width=0.30\textwidth,keepaspectratio]{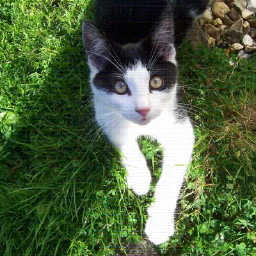}
    \includegraphics[width=0.30\textwidth,keepaspectratio]{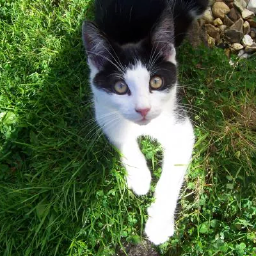}

    \includegraphics[width=0.30\textwidth,keepaspectratio]{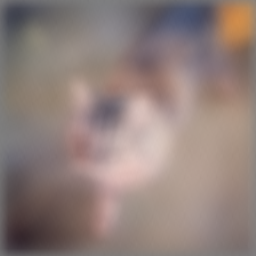}
    \includegraphics[width=0.30\textwidth,keepaspectratio]{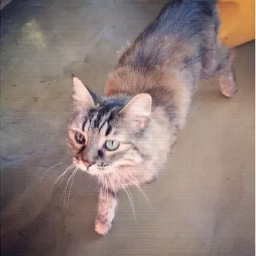}
    \includegraphics[width=0.30\textwidth,keepaspectratio]{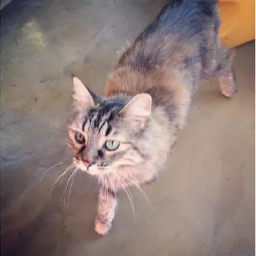}
    
    \caption{ MAP-GA(D,NP) based image deblurring (with $16\times 16$ uniform blur kernel) on LSUNCat256. Left to right: observed, recovered, original images. We choose $num\_steps=20$, $num\_iter=50$.}
    \label{fig:enter-label}
\end{figure*}%

\begin{figure*}
    \centering
    %\resizebox{.5\linewidth}{.25\textwidth}{%
    \includegraphics[width=0.30\textwidth,keepaspectratio]{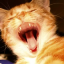}
    \includegraphics[width=0.30\textwidth,keepaspectratio]{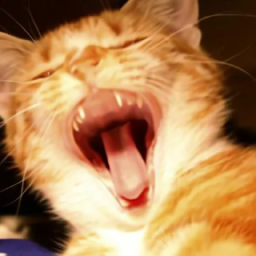}
    \includegraphics[width=0.30\textwidth,keepaspectratio]{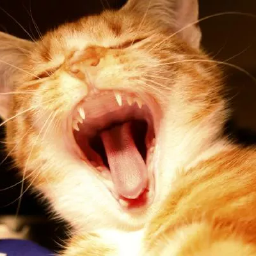}
    
    \includegraphics[width=0.30\textwidth,keepaspectratio]{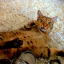}
    \includegraphics[width=0.30\textwidth,keepaspectratio]{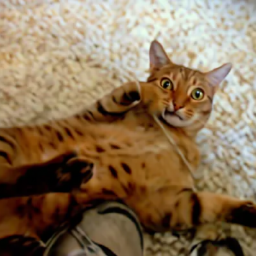}
    \includegraphics[width=0.30\textwidth,keepaspectratio]{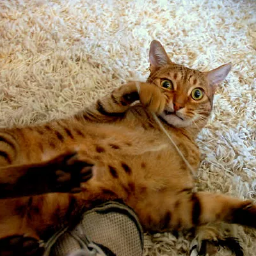}

    \includegraphics[width=0.30\textwidth,keepaspectratio]{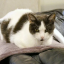}
    \includegraphics[width=0.30\textwidth,keepaspectratio]{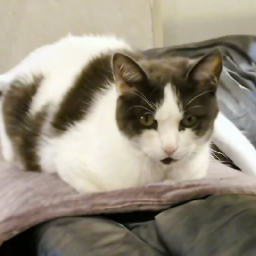}
    \includegraphics[width=0.30\textwidth,keepaspectratio]{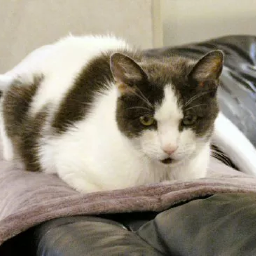}
    
    \caption{ MAP-GA(D,NP) based $4\times$ image super-resolution on LSUNCat256. Left to right: observed, recovered, original images. We choose $num\_steps=20$, $num\_iter=50$.}
    \label{fig:enter-label}
\end{figure*}%

\begin{figure*}
    \centering
    %\resizebox{.5\linewidth}{.25\textwidth}{%
    \includegraphics[width=0.24\textwidth,keepaspectratio]{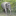}
    \includegraphics[width=0.24\textwidth,keepaspectratio]{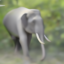}
    \includegraphics[width=0.24\textwidth,keepaspectratio]{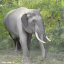}
    \includegraphics[width=0.24\textwidth,keepaspectratio]{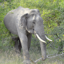}

    \includegraphics[width=0.24\textwidth,keepaspectratio]{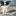}
    \includegraphics[width=0.24\textwidth,keepaspectratio]{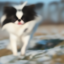}
    \includegraphics[width=0.24\textwidth,keepaspectratio]{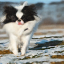}
    \includegraphics[width=0.24\textwidth,keepaspectratio]{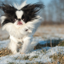}

    \includegraphics[width=0.24\textwidth,keepaspectratio]{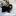}
    \includegraphics[width=0.24\textwidth,keepaspectratio]{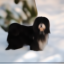}
    \includegraphics[width=0.24\textwidth,keepaspectratio]{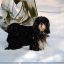}
    \includegraphics[width=0.24\textwidth,keepaspectratio]{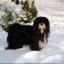}
    
    \caption{ $4\times$ image super-resolution on ImageNet64. Left to right: observed image, recovered image using MAP-GA(D,NP), recovered image using MAP-GA(NP), original image. We choose $num\_steps=20$, $num\_iter=50$ for both variants. Note that this $4\times$ downsampling corresponds to a severe degradation since the original image is of $64\times 64$ resolution}
    \label{fig:enter-label}
\end{figure*}%

\begin{figure*}
    \centering
    %\resizebox{.5\linewidth}{.25\textwidth}{%
    \includegraphics[width=0.25\textwidth,keepaspectratio]{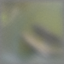}
    \includegraphics[width=0.25\textwidth,keepaspectratio]{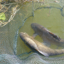}
    \includegraphics[width=0.25\textwidth,keepaspectratio]{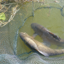}

    \includegraphics[width=0.25\textwidth,keepaspectratio]{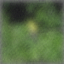}
    \includegraphics[width=0.25\textwidth,keepaspectratio]{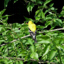}
    \includegraphics[width=0.25\textwidth,keepaspectratio]{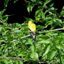}

    \includegraphics[width=0.25\textwidth,keepaspectratio]{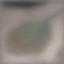}
    \includegraphics[width=0.25\textwidth,keepaspectratio]{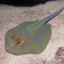}
    \includegraphics[width=0.25\textwidth,keepaspectratio]{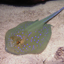}
    
    \caption{ MAP-GA(D,NP) based image deblurring (with $7\times 7$ uniform blur kernel) on ImageNet64. Left to right: observed, recovered, original images. We choose $num\_steps=20$, $num\_iter=50$.}
    \label{fig:enter-label}
\end{figure*}%

%%%%%%%%% REFERENCES
%{\small
%\bibliographystyle{ieee_fullname}
%\bibliography{egbib}
%}